\documentclass[review]{elsarticle}

\usepackage{multirow}
\usepackage[utf8]{inputenc}
\usepackage{color}
\usepackage{hyperref}

\newcommand{\remi}[2]{{\color{black}  #2}}
\newcommand{\cor}[1]{{\color{black}  #1}}

\usepackage{pstricks}
\usepackage{verbatim}
\usepackage{subfigure}
\usepackage{algorithmic}
\usepackage{algorithm}

\usepackage{amssymb}
\usepackage{amsmath}

 \usepackage{graphicx}     \topmargin -1.5cm         \oddsidemargin -0.04cm    \evensidemargin -0.04cm   \textwidth 16.59cm
\def\W{\ensuremath{\mathbf{W}}}

\def\w{\ensuremath{\mathbf{w}}}
\def\x{\ensuremath{\mathbf{x}}}

\def\y{\ensuremath{\mathbf{y}}}
\def\b{\ensuremath{\mathbf{b}}}
\def\u{\ensuremath{\mathbf{u}}}

\def\dbR{{\mathrm{I\hskip-2.2pt R}}}
\def\dbR{\mathbb{R}}

\def\R{\mathbb{R}}

\def\M{\mathbf{M}}
\def\I{\mathbf{I}}
\def\D{\mathbf{D}}

\begin{document}         
 \title{Mixed-norm Regularization for Brain Decoding}

\author[lagrange]{R. Flamary\corref{cor1}}
\ead{remi.flamary@unice.fr}
\author[gipsa]{N. Jrad}
\author[gipsa]{R. Phlypo}
\author[gipsa]{M. Congedo}
\author[litis]{A. Rakotomamonjy\corref{cor1}}
\ead{alain.rakoto@insa-rouen.fr}

\cortext[cor1]{Corresponding author}
\address[lagrange]{Laboratoire Lagrance, UMR7293, Université de Nice, 00006
Nice, France}
\address[litis]{LITIS, EA 4108 - INSA / Universit\'{e} de Rouen, 76000
Rouen, France}
\address[gipsa]{Gipsa Lab, Domaine Universitaire
BP 46, 38402 Saint Martin d'Hères cedex, France}

 \begin{abstract}
This work investigates the use of mixed-norm regularization for
 sensor selection in Event-Related Potential (ERP) based 
Brain-Computer Interfaces
(BCI). The classification problem is cast as a discriminative
optimization framework where sensor selection is induced through the
use of mixed-norms.
This framework is extended to the multi-task learning situation where
several similar classification tasks related  to  different
subjects are learned simultaneously.
In this case, multi-task
learning helps in leveraging data scarcity issue yielding
to more robust classifiers. For this purpose, we have introduced
a regularizer that induces both sensor selection and classifier
similarities. The different
regularization approaches are compared on three ERP datasets showing the
interest of mixed-norm regularization in terms of sensor
selection. The multi-task
approaches are evaluated
when a small number of learning examples are available yielding to
significant performance improvements especially for subjects performing poorly. 
                                          \end{abstract}


 \maketitle

\section{Introduction}
\label{sec:definitions}

Brain Computer Interfaces (BCI) are systems that
help disabled people communicating with their environment
through the use of  brain
signals \cite{dornhege2007toward}. 
At the present time, one of the most prominent BCI is based on
electroencephalography (EEG)
because of its low-cost, portability and its non-invasiveness.
 Among EEG based BCI, a paradigm of
interest is the one based
on event-related potentials (ERP) which are responses of the brain to 
some external stimuli. In this context, the innermost part of a BCI
is the pattern recognition stage which has to correctly recognize presence
of these ERPs. 
However, EEG signals are
blurred due to the diffusion of the skull and the
skin~\cite{nunez2006electric}. Furthermore, EEG recordings are highly
contaminated by noise of biological, instrumental and environmental
origins. For addressing these issues, advanced  signal processing and
machine learning 
techniques have been employed to learn ERP patterns from training EEG signals
leading to robust systems able to 
recognize the presence of these
events~\cite{blankertz_bciIII,rivet2010eeg,blankertz2010singletrial,muller99designing,gouypailler2010nonstationnary,salimi2008fusion}. 
Note that while some ERPs are used for generating BCI commands,
some others can be used for improving BCI efficiency.
Indeed, recent studies
have also tried to develop algorithms for automated recognition
of error-related potentials \cite{falkenstein1991effects}. These potentials are responses elicited when
a subject commits an error in a BCI task or observes an error
\cite{ferrez2007error,buttfield2006toward} and thus they
 can help in correcting errors or in providing feedbacks to
BCI user's.

In this context of automated recognition of event-related potentials
for BCI systems, reducing the number of EEG sensors is of primary
importance since it \remi{helps in learning robust classifiers by removing
irrelevant and noisy features. Furthermore, by doing so, one also
minimizes the implementation cost of the BCI (fewer EEG sensors, setup
speed, calibration time)}{reduces the implementation cost of the
BCI by minimizing the number of EEG sensor, and  speeding up
experimental setup and calibration time.}  For this purpose,  some studies have
proposed to choose relevant sensors
according to prior knowledge of brain functions. For instance, 
 sensors located above the motor cortex region are preferred for motor
 imagery tasks and while for visual Event Related Potential
(ERP), sensors located on the visual cortex are favored
\cite{krusienski2008toward}. Recent works have focused on
automatic sensor selection adapted to the specificity of a
subject~\cite{hoffman2008bayesian,lal2004support,yang2012channel,rivet2010eeg,cecotti2011robust,jrad2011sw}.
For instance, Rakotomamonjy et
al. ~\cite{rakotomamonjy08:_bci_compet_iii} performed a recursive
backward sensor selection using cross-validation classification
performances as an elimination criterion. 
Another approach for exploring subset sensors have been
proposed by \cite{yang2012channel}, it consists in using a genetic
algorithm for sensor selection coupled with an artificial neural
networks for prediction. Those methods has been proven
efficient but computationally demanding. 
A quicker way is to estimate the 
relevance of the sensors in terms of  Signal to Noise Ratio
(SNR)~\cite{rivet2010eeg} and to keep the most relevant ones. Note
that this approach does not optimize a discrimination criterion,
although the final aim is a classification task.
\cor{Recently, van Gerven et
al. \cite{van2009interpreting} proposed a graceful approach for
embedding sensor selection into a discriminative framework. They
performed sensor selection and learn a decision function by solving a
unique optimization problem. In their framework, a logistic regression
classifier is learned and the group-lasso regularization, also known as
$\ell_1-\ell_2$ mixed-norm, is used to promote sensor selection. They
have also investigated the use of this groupwise regularization for
frequency band selection  and their applications to transfer learning. The same idea has been explored by
Tomioka et al. \cite{tomioka2010regularized} which also considered groupwise
regularization for classifying EEG signals.
In this work, we go beyond these studies  by providing an in-depth study of
the use of mixed-norms for sensor selection in a single subject
setting and  by discussing the utility of mixed-norms when learning
decision functions for multiple subjects simultaneously.
}

Our first contribution addresses the problem of robust sensor
selection  embedded into a discriminative framework. We broaden
the analysis of van Gerven et al. \cite{van2009interpreting} by considering 
regularizers which forms are  $\ell_1-\ell_q$ mixed-norms,
with $(1\leq q \leq2)$, as well as adaptive mixed-norms,
 so as to promote sparsity among group of features or
sensors. In addition to  providing a sparse and accurate
sensor selection, mixed-norm regularization has several
advantages. First, sensor selection is cast into an elegant 
discriminative framework, using for instance a large margin paradigm,  which does not require any additional hyper-parameter to be optimized. Secondly, 
since sensor selection is jointly learned with the
classifier by optimizing an ``all-in-one'' problem, selected
sensors are directed to the goal of discriminating relevant
EEG patterns. 
Hence, mixed-norm regularization helps locating sensors which are
relevant for an optimal classification performance.

A common drawback of all the aforementioned sensor selection
techniques is that selected set of sensors may vary, more or less
substantially, from subject to subject. This variability, \remi{ increases
inversely proportional with the number of available trials per
subject}{  is due partly to subject specific differences  and partly to
acquisition noise and limited number of training examples}. In such a case, selecting a robust subset of sensors
may  become a complex problem. Addressing this issue is the point of
our second contribution. We propose a Multi-Task Learning (MTL)
framework that helps in learning robust classifiers able to
cope with the scarcity of learning examples.
MTL is one way of achieving
inductive transfer between tasks. The goal of inductive transfer is to
leverage additional sources of information to improve the performance
of learning on the current task.  The main hypothesis underlying MTL
is that tasks are related in some ways. In most cases, this
relatedness is translated into a prior knowledge, \emph{e.g} a
regularization term, that a machine learning algorithm can take
advantage of. For instance, regularization terms may promote
similarity between all the
tasks~\cite{evgeniou04:_regul_multi_task_learn}, or enforce classifier
parameters to lie in a low dimensional linear
subspace~\cite{argyriou_multitasklong}, or to jointly select  the
relevant features~\cite{rakotomamonjy2011lplqpenalty}. MTL has been
proven efficient for motor imagery in~\cite{alamgir2009multi} where
several classifiers were learned simultaneously from several BCI
subject datasets.\cor{~Our second contribution is thus focused on the problem
of   performing sensor selection and learning
robust classifiers  through the use of an MTL mixed-norm
regularization framework. We propose a novel regularizer  promoting sensor
selection  and similarity between classifiers. 
By doing so, our goal is then to yield  sensor selection
and  robust classifiers that are able to overcome the data scarcity problem
by sharing information between the different classifiers to be learned.}

The paper is organized as follows. The first part of the paper presents 
the discriminative
framework and the different regularization terms we have considered for
channel selection and multi-task learning. The
second part is devoted to the description of the datasets, the
preprocessing steps applied to each of them and
 the results achieved in terms of performances and sensor selection.
\cor{In order to promote reproducible research, the code needed for
  generating the
results in this paper is available of the author's
website \footnote{URL:
  \url{http://remi.flamary.com/soft/soft-gsvm.html}}.}

\section{Learning framework}
\label{sec:single-task-sensor}

In this section, we introduce our mixed-norm 
regularization framework  that can
be used to perform sensor selection in
a single task or in a transfer learning setting.

\subsection{Channel selection in a single task learning setting}
\label{sec:single-task-learning}
Typically in BCI problems, one wants to learn
a classifier that is able to predict the class of
some EEG trials, from a set of learning examples.   
We denoted as $\{\x_i,y_i\}_{i\in\{1\dots n\}}$  the learning set such that
$\x_i\in\dbR^d$ is a trial and 
$y_i\in\{-1,1\}$ is its corresponding class, usually related
to the absence or presence of an event-related potential. 
In most cases, a trial $\x_i$ is extracted from a multidimensional signal and 
thus is characterized by  $r$ features for each of the $p$ sensors, 
leading to a dimensionality $d=r\times
p$.
Our aim is to learn, for a single subject, a
linear classifier $f$ that will predict the class of a
trial $\x\in \dbR^d$, by looking at the sign of the function $f(\cdot)$ defined as:
\begin{equation}
  \label{eq:predict}
  f(\x)=\x^T\w+b
\end{equation}
with $\w\in\dbR^d$ the normal vector to the separating hyperplane and $b\in \dbR$ a bias term. 
Parameters of this function are learned by solving the optimization problem:
\begin{equation}
  \label{eq:disc_framework}
  \min_{\w,b}\quad \sum_i^n L_o(\y_i,\x_i^T\w+b)+ \lambda\Omega(\w)
\end{equation}
where $L_o$ is a loss function that measures the discrepancy between 
actual and  predicted labels,
$\Omega(\cdot)$  a regularization term that expresses some
prior knowledge about the learning problem and $\lambda$ a
parameter that balances both terms.  
In this work, we choose $L_o$ to be the
squared hinge loss
$L_o(y,\hat y)=max(0,1-y\hat y)^2$,  thus promoting a large margin classifier.

\subsubsection{Regularization terms}
\label{sec:regularization-terms}

We now discuss different regularization terms that may be
used for single task learning along with their significances in terms of
channel selection.

\paragraph{$\ell_2$ norm}
\label{sec:ell_2}
The first regularization term that comes to mind is the standard
squared $\ell_2$ norm regularization:
\begin{equation}
  \label{eq:l2}
  \Omega_2(\w)=\frac{1}{2}||\w||_2^2
\end{equation}
where $||\cdot||_2$ is the Euclidean norm.
This is the common regularization term used for SVMs and it
will be considered in our experiments as the baseline approach. \cor{
Intuitively, this regularizer tends to downweight the amplitude
of each component of $\w$ leading to a better control of 
the margin width of our large-margin classifier and thus it helps in  
reducing overfitting. }

\paragraph{$\ell_1$ norm}
\label{sec:ell_1-norm}
When only few of the features are discriminative for a
classification task, a common way to select the relevant ones is
to use an $\ell_1$ norm of the form
\begin{equation}
  \label{eq:l2}
  \Omega_1(\w)=\sum_{i=1}^d|w_i|
\end{equation}
as a regularizer \cite{bach2011convex}. Owing
to its mathematical properties (non-differentiability at $0$), 
unlike the $\ell_2$ norm, this regularization term promotes sparsity,
which means that at optimality of  problem (\ref{eq:disc_framework}), some components of $\w$ are exactly $0$. \cor{In a Bayesian framework, the   $\ell_1$ norm is related to the use of prior on $\w$ that forces its component
to vanish \cite{van2009interpreting}. This is typically obtained by means of Laplacian prior over the weight.}
However, $\ell_1$ norm  ignores the structure
of the features  (which may be grouped by sensors) \cor{since each component of $\w$ is treated independently to the others} yielding
thus to feature selection but not to sensor selection.

\begin{figure}[t]
  \centering
  \includegraphics[width=8cm]{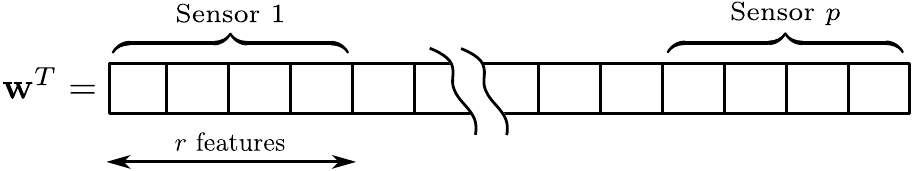}\\
\vspace{.4cm}
    \includegraphics[width=8cm]{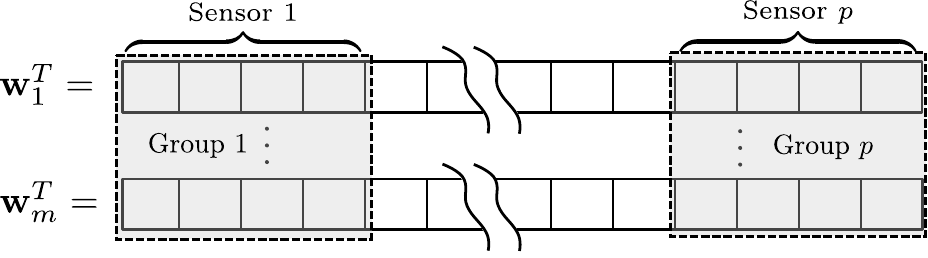}

  \caption{Examples of feature grouping for (top) single task and (bottom) multiple task learning. }
  \label{fig:groupfeat}
\end{figure}

\paragraph{$\ell_1-\ell_q$ mixed-norm}
\label{sec:ell_1-ell_2-mixed}
A way to take into account the fact
that  features are structured, is to use a mixed-norm that will group them and regularize them together. Here, we consider mixed-norm of the form 
\begin{equation}
  \label{eq:mixednorm}
  \Omega_{1-q}(\w)=\sum_{g\in\mathcal{G}}||\w_{g}||_q
\end{equation}
with $1\leq q
\leq 2$ and $\mathcal{G}$ being a partition of the set $\{1,\cdots,d\}$. 
\cor{Intuitively, this $\ell_1-\ell_q$ mixed-norm can be interpreted
as  an $\ell_1$ norm applied to
the vector containing the 
 $\ell_q$ norm of each group of features. It  promotes sparsity on
each $\w_g$ norm and consequently on the  $\w_g$ components
as well.}
For our BCI problem, a natural choice for $\mathcal{G}$ is to group the
features by sensors yielding thus to  $p$ groups (one per sensor) of $r$ features as reported in Figure \ref{fig:groupfeat}. 
Note that unlike the $\ell_1-\ell_2$ norm as used by van Gerven et al. \cite{van2009interpreting} and 
Tomioka et al. \cite{tomioka2010regularized}, the
use of an inner  $\ell_q$ norm leads to more flexibility as it spans from the
$\ell_1-\ell_1$ (equivalent to the $\ell_1$-norm and leading thus to  unstructured feature selection) to the $\ell_1-\ell_2$ \cor{which strongly ties together the
components of a group. Examples of the use of $\ell_q$ norm and 
mixed-norm regularizations
in other biomedical contexts can be found for instance in \cite{rahimi13,liu13:_non_negat_mixed_norm_convex}.}  

\paragraph{Adaptive $\ell_1-\ell_q$}
\label{sec:adaptive-ell_1-ell_2}

The  $\ell_1$ and $\ell_1-\ell_q$ norms described above, are well-known to lead to grouped feature selection. 
However,  they are also known, to lead to poor
statistical properties (at least when used with a square
loss function)~\cite{bach_consistencymkl}. \cor{For instance, they
are known to have consistency issue in the sense that, even with an 
arbitrarily large  number
of training examples, these norms may be unable to select
the true subset of features. In practice,
this means that when used in Equation (\ref{eq:disc_framework}),
the optimal weight vector $\w$ will tend to 
over-estimate the number of relevant sensors.}
These issues can be addressed by   considering an adaptive
$\ell_1-\ell_q$ mixed-norm of the form \cite{zou_adaptive_lasso_2006,bach_consistencymkl}:
 \begin{equation}
  \label{eq:mixednorm}
  \Omega_{a:1-q}(\w)=\sum_{g\in\mathcal{G}}\beta_g||\w_{g}||_q
\end{equation}
where the weights $\beta_g$ are selected so as to enhance the sparsity
pattern of $\w$. In our
experiments, we obtain them by first solving  the $\ell_1-\ell_q$
problem with $\beta_g=1$, which outputs
an optimal parameter $\w^*$, and  by finally defining $\beta_g=1/||\w^*_{g}||_q$.
Then, solving the weighted $\ell_1-\ell_q$ problem yields
an optimal solution with  increased sparsity pattern compared
to $\w^*$ since the $\beta_g$ augments the penalization \cor{of
groups with  norm $\|\w^*_{g}\|_q$ smaller than $1$.} 

\subsubsection{Algorithms}
\label{sec:algorithms}

Let us now discuss how  problem (\ref{eq:disc_framework}) is solved when
one of these regularizers is in play.  

Using the $\ell_2$ norm  regularization makes the problem  differentiable. Hence a  first or second-order descent based algorithm can be considered~\cite{chap_primal}. 

Because the other regularizers are not differentiable, \cor{we have deployed
an algorithm \cite{combettes2011proximal} tailored for minimizing
objective function of the form $f_1(\w)+f_2(\w)$ with $f_1$ a smooth and differentiable convex function with Lipschitz constant $L$ and  $f_2$ a continuous and convex non-differentiable function 
having a simple
proximal operator, 
\emph{i.e.} a closed-form or an easy-to-compute solution of the problem:
\begin{equation}
\text{prox}_{f_2}(\mathbf{v}):=\text{argmin}_\u \frac{1}{2}\|\mathbf{v}-\u\|_2^2+f_2(\u)\label{eq:1}
\end{equation}
Such an algorithm, known as forward-backward splitting \cite{combettes2011proximal}~is simply based on the following iterative approach,
\begin{equation}\label{eq:proximal}
\w^{k+1}= \text{prox}_{\frac{1}{\gamma} f_2}(\w^{k} - \gamma \nabla_{\w} f_1(\w^{k}))
\end{equation}
with $\gamma$ being a stepsize in the gradient descent. This algorithm can be easily derived by considering,
 instead of directly minimizing $f_1(\w) + f_2(\w)$, an iterative scheme
which at each iteration replace  $f_1$ with a quadratic approximation of
$f_1(\cdot)$ in the neighborhood of $\w^k$. Hence,
$\w^{k+1}$ is the minimizer of ~:
$$
f_1(\w^k) + \langle \nabla_\w f_1(\w^k), \w-\w^k \rangle + \frac{\gamma}{2}
\|\w-\w^k\|_2^2 + f_2(\w)
$$  
which closed-form is given in Equation (\ref{eq:proximal}).
This algorithm is known to converge towards
a minimizer of $f_1(\w) + f_2(\w)$ under some weak conditions on the
stepsize \cite{combettes2011proximal}, which is satisfied by
choosing for instance $\gamma = \frac{1}{L}$. 
We can note that the algorithm defined in Equation (\ref{eq:proximal}) has the same flavor as a projected
gradient algorithm which first, takes a gradient step, and then 
``projects'' back the solution owing to the proximal operator. More
details can also be found in \cite{beck09:_fast_iterat_shrin_thres_algor}.
}

\cor{
For our problem (\ref{eq:disc_framework}), we choose $f_1(\w)$ to be the squared hinge loss and $f_2(\w)$ the non-smooth regularizer.  
The square hinge loss is indeed  gradient Lipschitz with a constant
$L$ being $2 \sum_{i=1} \|\x_i\|_2^2$. Proof of this statement
is available in \remi{the supplementary material \cite{flamary12:_appen_mixed_brain_decod}.}{ Appendix \ref{sec:append_lip}.}
Proximal operators of the
$\ell_1$ and the $\ell_1-\ell_2$ regularization term can be 
easily shown to be the soft-thresholding and the block-soft thresholding
operator \cite{bach2011convex}. The general $\ell_1-\ell_q$ norm does not admit a closed-form solution, but 
its proximal operator can be simply computed 
by means of an iterative algorithm~\cite{rakotomamonjy2011lplqpenalty}. 
More details on these proximal operators are also available in
\remi{}{ Appendix \ref{sec:append_prox}.}
}

\subsection{ Channel selection and transfer learning in multiple
task setting}
\label{sec:multiple-task-learning}

We now address the problem of channel selection in cases
where  training examples for several subjects are at our disposal.
We have claimed that in such a situation, it would be 
benefitial to learn the decision functions related to all subjects
simultaneously, while inducing selected channels to be alike
for all subjects, as well as inducing decision function parameters
to be related in some sense.  These two hypotheses make reasonable sense
since brain regions related to the appearance of a given ERP are expected
to be  somewhat location-invariant across subjects. 
For solving this problem, we apply a machine learning paradigm, known
as multi-task learning, where in our case, each task is related to the
decision function of a given subject and where the regularizer should
reflect the above-described prior knowledge on the problem.  Given 
$m$ subjects, the
resulting optimization problem boils down to be
\begin{equation}
  \label{eq:mtlframework}
  \min_{\W,\b}\quad\sum_t^m \sum_{i=1}^{n_t} L(y_{i,t},\x_{i,t}^T\w_t+\b_t)+
  \Omega_{\text{mtl}}(\W)
\end{equation}
with $\{\x_{i,t},y_{i,t}\}_{i\in\{1\dots n_t\}}$ being the training
examples related to each task $t\in{1\dots m}$, $(\w_t,\b_t)$ being
the classifier parameters for task $t$ and $\W=[\w_1\dots
\w_m]\in\dbR^{d \times m}$ being a matrix concatenating all vectors
$\{\w_t\}$. \cor{Note that the multi-task learning framework applied
  to single EEG trial classification have already been investigated by
  van Gerven et al. \cite{van2009interpreting}. The main contribution
  we bring compared to their works is the use of regularizer that
  explicitly induces all subject classifiers to be similar to an
  average one, in addition to a regularizer that enforces selected
  channels to be the same for all subjects.  The intuition behind this
  point is : we believe that since the classification tasks we are
  dealing with, are similar for all subjects and all related
  to the same BCI paradigm, selected channels and
  classifier parameters should not differ that much from subject to
  subject.  We also think that inducing task parameters to be similar
  may be more important than enforcing  selected channels to be similar when
  the number of training examples is small since it helps in reducing
  overfitting. } For this purpose, we have proposed a novel
regularization term of the form~:
\begin{equation}
  \label{eq:regterm_mtl}
  \Omega_{\text{mtl}}(\W)=\lambda_r \sum_{g\in\mathcal{G'}}||\W_{g}||_2+\lambda_s\sum_{t=1}^m||\w_t-\hat\w||^2_2
\end{equation}
where $\hat\w=\frac{1}{m}\sum_t \w_t$ is the average classifier across
tasks and $\mathcal{G'}$ contains non-overlapping groups of components
from matrix $\W$. 
The first term in Equation (\ref{eq:regterm_mtl}) is a mixed-norm
term that promotes group regularization. In this work, we
defined groups in $\mathcal{G'}$ based on the sensors, which
means that  all the features across subject related to a given sensor are in
the same group $g$, leading to $p$  groups of $r\times m$ feature,
as depicted in Figure \ref{fig:groupfeat}.
 The second term is a similarity promoting term as introduced in
  Evgeniou et al. \cite{evgeniou04:_regul_multi_task_learn}. It
can be interpreted  as a term enforcing the
minimization of the classifier's parameter variance. 
 In other
words, it promotes classifiers to be similar to the average
one, and it helps  improving performances when the number of
learning examples for each task is limited, by reducing over-fitting.
Note that $\lambda_r$ and $\lambda_s$ respectively control the sparsity induced
by the first term and the similarity induced by the second one.
\cor{Hence, when setting $\lambda_s=0$, the regularizer given in 
Equation (\ref{eq:regterm_mtl}) boils down to be similar to the
one used by van Gerven et al. \cite{van2009interpreting}.
Note that in practice  $\lambda_r$ and $\lambda_s$ are selected by means
of a nested cross-validation which aims at classification accuracy.
Thus, it may occur that classifier similarity is preferred
over  sensor selection leading to robust classifiers which still
use most of the sensors.
}

Similarly to the single task optimization framework given in 
Equation (\ref{eq:disc_framework}), the objective function
for problem \eqref{eq:mtlframework} can be expressed as a \cor{sum of gradient
Lipschitz continuous term
$f_1(\W)=\sum_{t,i}^{m,n}L(\cdot)+\lambda_s\sum_{t=1}^m||\w_t-\hat\w||^2_2$}
and a non-differentiable term
$f_2(\W)=\lambda_r \sum_{g\in\mathcal{G'}}||\W_{g}||_2$  having a
closed-form proximal operator (see Appendix \ref{sec:append_lip2}). 
Hence, we have again considered a forward-backward
splitting algorithm which iterates are given in Equation (\ref{eq:proximal}).

\section{Numerical experiments}

We now present how these novel approaches perform 
on different  BCI problems. Before delving into the details
of the results, we introduce the simulated and real  datasets.

\subsection{Experimental Data}

We have first evaluated the proposed approaches on a simple simulated
P300 dataset  generated as follows. A P300 wave is extracted using the
grand
average of a single subject data from the EPFL dataset described in the
following.
We generate
$11000$ simulated examples
with 8 discriminative channels containing the P300 out of 16
channels for positive examples. A  Gaussian noise of
standard deviation $0.2$ is added to all signals making
the dataset more realistic. $1000$ of these examples have 
been used for training.

The first real P300 dataset we used is the EPFL
dataset,
 based on eight subjects performing P300 related 
tasks~\cite{hoffmann08:_effic_p300_based_brain_comput}. The subjects
were asked to
focus on one of the 3$\times$2=6 images on the screen while the one of the images
is flashed at random. The EEG signals
were acquired from 32 channels,
sampled
at $1024$ Hz and
4 recording sessions per subject have been realized.  Signals are 
pre-processed exactly according to the steps described in
\cite{hoffmann08:_effic_p300_based_brain_comput} : a  $[1,8]$~Hz bandpass
 Butterworth filter of order 3 is applied to all
 signals followed by a downsampling. Hence, for each
trial (training example), we have 8 time-sample features
 per channel corresponding to a 1000 ms time-window after
 stimulus,  which leads to $256$ features for all
channels (32$\times$8=256 features). On the overall, the training set of a given
subject is composed of about 3000 trials.

Another P300 dataset, recorded by the Neuroimaging Laboratory
of Universidad 
Autónoma Metropolitana (UAM, Mexico)
\cite{ledesma2010open},
has also been utilized.
The data  have been obtained from  30 subjects
performing P300 spelling tasks on a 6$\times$6 virtual keyboard.
Signals are recorded over  10 channels 
leading thus to a very challenging
dataset for sensor selection, as there are just few sensors left to
select.
For this dataset, we only use the first 3
sessions in order to have the same number of trials for all subjects
($\approx$4000 samples).
The EEG signals have been pre-processed according to the
following steps~:  a  $[2,20]$ Hz Chebychef
bandpass filter of order 5 is first applied 
 followed by a decimation, resulting in a post-stimulus time-window of
 31 samples per channels. Hence, each trial is 
composed of $310$ (10$\times$31) features .

We have also studied the effectiveness of our methods on 
an Error Related Potential (ErrP) dataset that has been recorded in the GIPSA
Lab. The subjects were asked to memorize the position of 2 to 9
digits and to remind the position of one of these
digits, operation has been repeated $72$ times for each subject.
The signal following the visualization of the result
(correct/error on the memorized position) was recorded from 31 electrodes 
 and sampled at $512$ Hz. Similarly to Jrad et al. 
\cite{jrad2011sw}, a $[1,10]$~Hz Butterworth filter of order 4 and a downsampling has been applied to all channel signals. Finally,  a
time window of 1000ms is considered as a trial (training example)
with a
dimensionality of   $16\times 31=496$.

\subsection{Evaluation criterion, methods and experimental protocol}
\label{sec:methods-evaluation}

We have compared several regularizers that induce
feature/channel selection embedded in the learning algorithm,
in a single subject learning setting as
defined in Equation (\ref{eq:disc_framework}). 
\remi{
\cor{Because ERP classification datasets are highly
imbalanced, we use the Area Under the Roc Curve
(AUC) as a performance measure. This measure is an estimate of the
probability for a positive class to have a higher score than a
negative class.
}}{The performance measure commonly used in BCI Competitions
\cite{blankertz_bciIII} is the Area Under the Roc Curve
(AUC). This measure is an estimate of the probability
for a positive class to have a higher score than a
negative class. It makes particularly sense to use AUC when evaluating a P300 speller   as the letter in the keyboard is usually chosen by comparing score returned by the
  classifier for every column or line. In addition,  AUC does not
  depend on the proportion of positive/negative examples in the data
  which makes it more robust than classification error rate.}
Our baseline
algorithm is an SVM, which uses an $\ell_2$ regularizer and thus
does not perform any selection. Using an $\ell_1$ regularizer yields
a classifier which embeds feature selection, denoted as SVM-1 in the sequel.   
Three mixed-norm regularizers inducing sensor selection have also been 
considered : an  $\ell_1-\ell_2$ denoted as GSVM-2, and $\ell_1-\ell_q$
referred as GSVM-q, with $q$ being selected
  in the  set $\{1,1.2,\dots,1.8,2\}$) by a nested cross-validation stage,
and adaptive $\ell_1-\ell_q$ norm,
with $q=2$
denoted as GSVM-a.

For the multi-task learning setting, two MTL methods were compared to two
baseline approaches which use all features, namely 
a method that treats each tasks separately by learning one SVM per
task (SVM), and a method denoted as SVM-Full, which
on the contrary learns an unique SVM from all subject datasets. 
The two MTL methods are respectively a MTL as described in Equation
(\ref{eq:mtlframework}), denoted as MGSVM-2s and the same MTL but without
similarity-promoting regularization term, which actually means that
we set $\lambda_s=0$, indicated as MGSVM-2.  
For these approaches, performances are evaluated as the average
AUC of the decision functions over all the subjects.

The experimental setup is described in the following. For each subject, the dataset is randomly
split into a training set of $n=1000$ trials and a test set containing
the rest of the trials.  The regularization parameter
$\lambda$  has been selected from a log-spaced grid ($[10^{-3},10^1]$)
according to a nested $3$-fold
cross-validation step on the training set.
When necessary, the selection of $q$ is also included in this CV procedure. 
Finally, the selected value of $\lambda$  is used to learn  a
classifier on the training examples and performances are evaluated on the
independent test set. We run this procedure 10 times for
every subject and report average performances. 
A Wilcoxon signed-rank test, \cor{which takes ties
into account} is
used to evaluate the statistical difference of the mean
performances of all methods compared to the baseline SVM.
 We believe that such a test is more
appropriate for comparing methods than merely looking at 
the standard deviation due to the high inter-subject variability in BCI
problems.

\subsection{Results and discussions}
\label{sec:results}

We now present the results we achieved on the above-described datasets.

\subsubsection{Simulated dataset}
\label{sec:simulated-dataset}

\begin{table}[t]
  \centering
  \begin{tabular}{|l|c|c|c|c|}\hline
Methods & Avg AUC & AUC p-val & Avg Sel & F-measure\\
\hline
SVM       & 79.79 & - & 100.00  & 66.67 \\ 
GSVM-1   & 79.32 & 0.027 & 98.75  & 67.25 \\ 
GSVM-2   & \textbf{80.96} & 0.004 & 62.50  & 89.72 \\ 
GSVM-p  & 80.74 & 0.020 & 63.12  & 89.40 \\ 
GSVM-a& 80.51 & 0.014 & \textbf{45.62}  & \textbf{93.98} \\ 
\hline
  \end{tabular}
  \caption{Performance results on the simulated dataset :
the average   performance in
     AUC (in $\%
$), the average percent of selected sensors (Sel)
and the F-measure of the selected channels (in $\%$).
Best results for each performance measure are in bold. The p-value refers
to the one of a Wilcoxon signrank test with SVM as a baseline.}
  \label{tab:perfsimulated}
\end{table}

Average (over 10 runs) performance of the different regularizers on
the simulated dataset
are reported in Table \ref{tab:perfsimulated} through  AUC, 
sensor selection rate and F-measure. 
This latter criterion measures the relevance of the selected channels
compared to the true relevant ones. F-measure is formally
defined as
$$
\text{F-measure} = 2 \frac{|\mathcal{C}\cap \mathcal{C^*}|}{|\mathcal{C^*}| + |\mathcal{C}|}
$$  
where $\mathcal{C}$ and  $\mathcal{C}^*$ are respectively the set of selected
channels and true relevant channels and $|\cdot|$ here denotes the cardinality of a
set. Note that if the selected channels are all the relevant ones, then the
F-measure is equal to one. 
Most of the
approaches provide similar AUC performances. We can although
highlight that  group-regularization  approaches (GSVM-2,GSVM-p, GSVM-a)
drastically reduce the number of selected channels since only   
$62\%$ and $45\%$ of
the sensors are selected. A clear advantage goes to  the adaptive regularization that is both sparser and is more capable of retrieving the 
true relevant channels.

\subsubsection{P300 Datasets}
\label{sec:p300-datasets}

\begin{table*}[t]
  \centering
  \begin{tabular}{|l||c|c|c||c|c|c||c|c|c|}\hline
Datasets & \multicolumn{3}{|c||}{EPFL Dataset (8 Sub., 32 Ch.)}&
\multicolumn{3}{|c||}{UAM Dataset (30 Sub., 10 Ch.)}&\multicolumn{3}{|c|}{ErrP
  Dataset (8 Sub., 32 Ch)}\\\hline\hline
Methods   &Avg AUC  & Avg Sel &p-value &       Avg AUC  &Avg Sel
&p-value  &   Avg AUC  & Avg Sel &p-value \\ \hline

SVM       & 80.35 & 100.00& - & 84.47 & 100.00 & -& {76.96} & 100.00 &-  \\
SVM-1     & 79.88 & 87.66 & 0.15 & 84.45 & 96.27 & 0.5577& 68.84 & 45.85 & 0.3125 \\
GSVM-2    & \textbf{80.53} & 78.24 & 0.31 & \textbf{84.94} & {88.77} & 0.0001 &\textbf{ 77.29} & {29.84} & 0.5469 \\
GSVM-p    & {80.38} & {77.81} & 0.74 & \textbf{84.94} & 90.80 & 0.0001 & 76.84 & 37.18 & 0.7422 \\
GSVM-a    & 79.01 & \textbf{26.60} & 0.01 & 84.12 &\textbf{45.07}& 0.1109& 67.25 & \textbf{7.14}
& 0.1484 \\\hline
  \end{tabular}
  \caption{Performance results for the 3 datasets the average
    performance (over subjects) in
     AUC (in $\%
$), the average percent of selected sensors (Sel) and the  p-value of
     the Wilcoxon signrank test for the AUC when compared to the baseline SVM's one. Best performing
algorithms for each performance measure are in bold.}
  \label{tab:perfresults}
\end{table*}

Results for these datasets are reported in
Table \ref{tab:perfresults}.
For the EPFL dataset, all methods achieve performances that 
are not statistically different. However, we note
that GSVM-2 leads to sensor selection (80\% of sensor
selected) while GSVM-a yields  to  classifiers that, on
average, use $26\%$ of the
sensors  at the cost of a slight loss in performances (1.5\%
AUC). 

Results for the UAM dataset follow the same trend in term of
sensor selection  but 
we also observe that the mixed-norm
regularizers  yield to increased performances. GSVM-2 performs 
statistically better than SVM although most of the sensors  (9 out of 10)
have been kept in the model. This
shows that even if few channels have been removed, the
group-regularization improves performances by bringing sensor prior knowledge to the problem.
We also notice that GSVM-a performance is statistically equivalent to the 
 baseline SVM one while using only half of the sensors and GSVM-p
consistently gives
similar  results to GSVM-2. 

To summarize, concerning the performances of the different mixed-norm
regularization, we outline that on one hand, GSVM-2 is at worst,
equivalent to the baseline SVM while achieving sensor selection and
on the other hand GSVM-a yields to a substantial channel selection at
the expense of a slight loss of performances.

\begin{figure}[t]
  \centering
  \includegraphics[width=0.9\columnwidth]{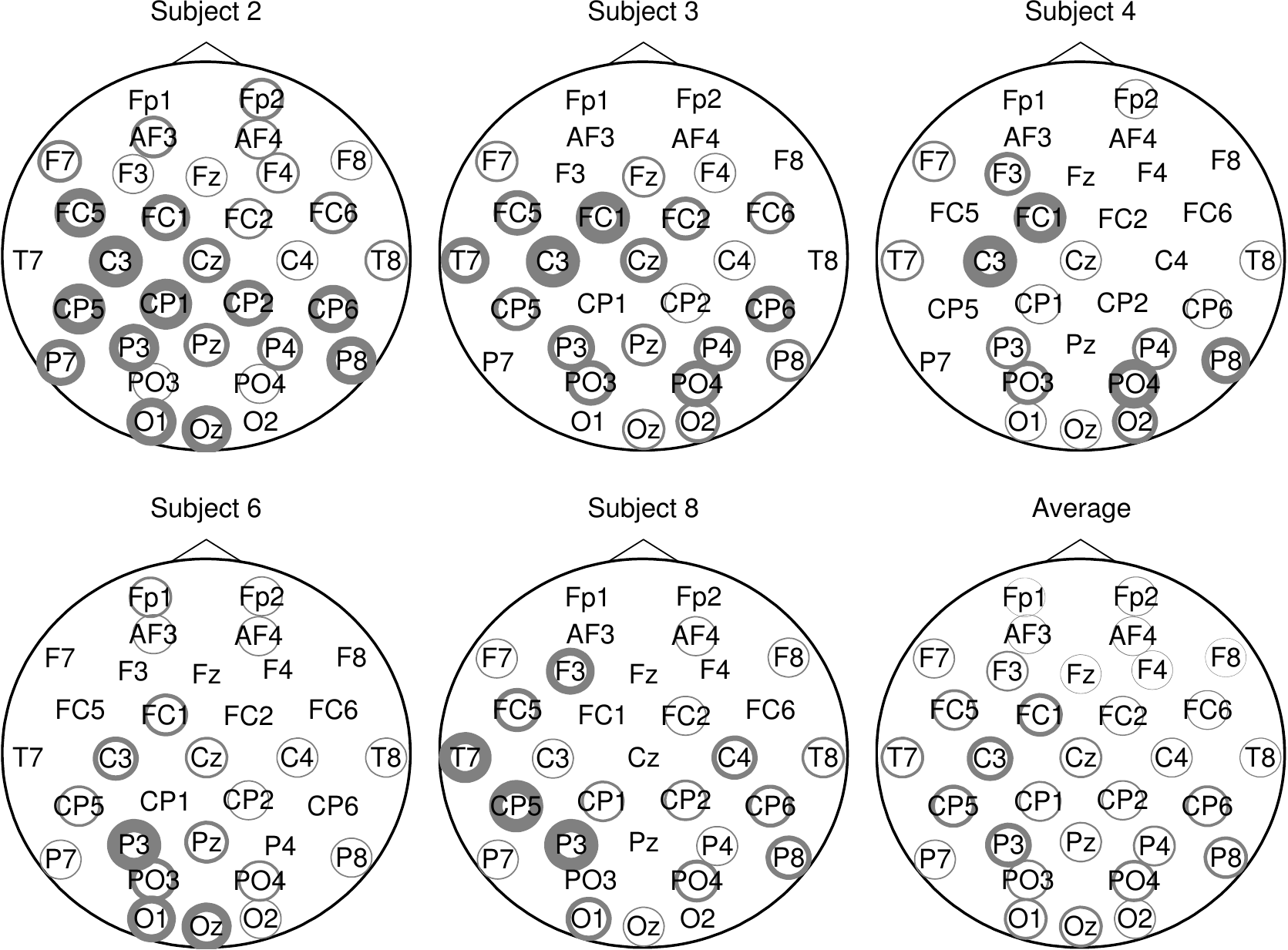}
  \caption{Selected sensors for the EPFL dataset. The line width of the
    circle is proportional to the number of times the sensor is
    selected for different splits. No circle means that the sensor has
    never been selected.}
  \label{fig:sel_sensor_epfl}
\end{figure}

\begin{figure}[t]
  \centering
  \includegraphics[width=0.9\columnwidth]{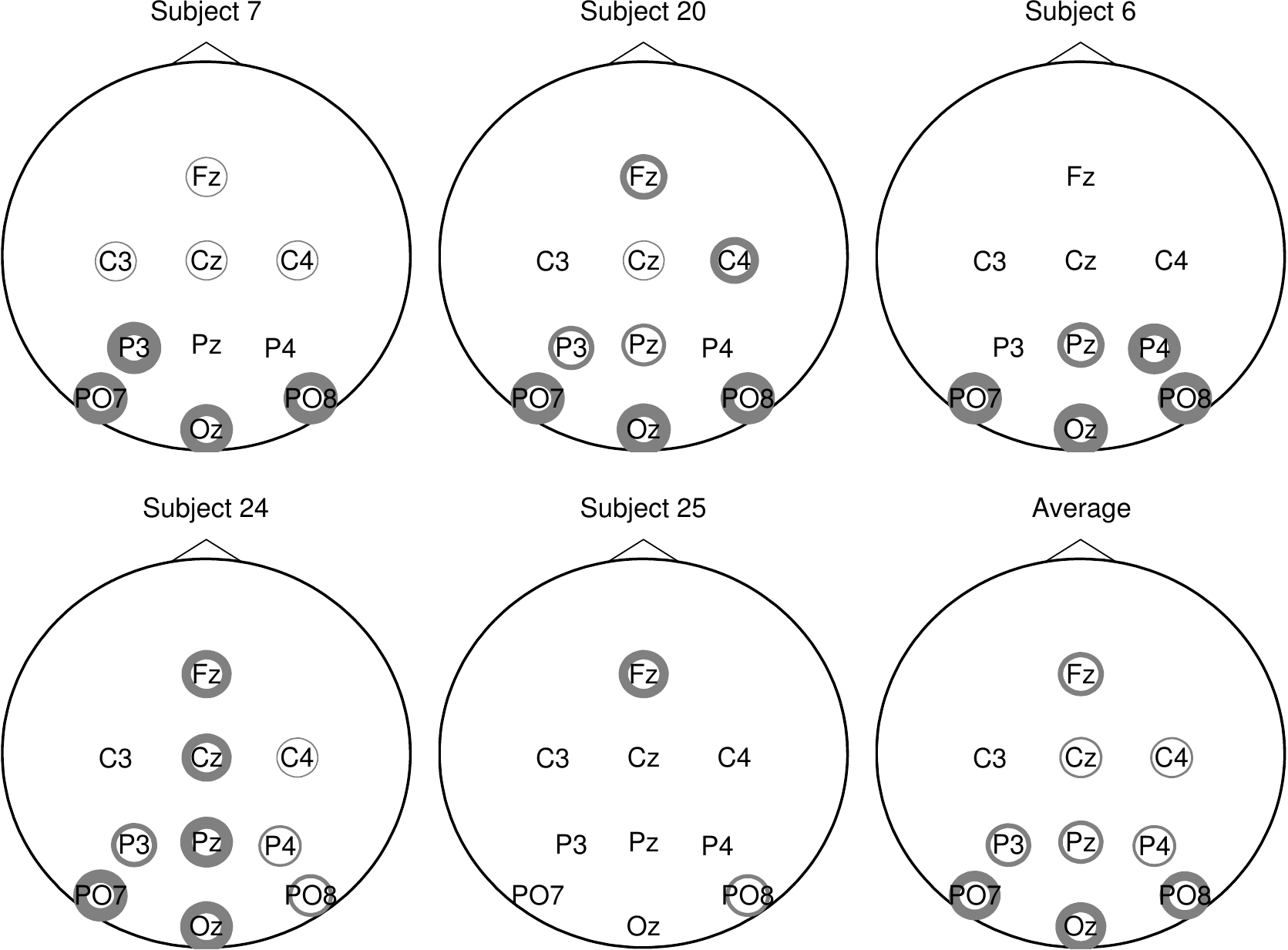}
  \caption{Selected sensors for the UAM dataset. The line width of the
    circle is proportional to the number of times the sensor is
    selected for different splits. No circle means that the sensor has
    never been selected.}
  \label{fig:sel_sensor_uam}
\end{figure}

A visualization of the electrodes selected by \mbox{GSVM-a} can be
seen in Figure
\ref{fig:sel_sensor_epfl} for the EPFL dataset and in Figure
\ref{fig:sel_sensor_uam} for the UAM dataset. Interestingly, we observe  
that for the EPFL dataset, the
selected channels are highly
dependent on the subject. The most recurring ones are : FC1
C3 T7 CP5 P3 PO3 PO4 Pz and the electrodes located above visual cortex  O1,Oz and
O2. We see sensors from the occipital area
that are known to be relevant~\cite{krusienski2008toward} for 
P300 recognition, but
sensors such as T7 and C3, from other
 brain regions are also frequently selected. These results
are however consistent with those presented
in the recent literature~\cite{rivet2010eeg,rakotomamonjy08:_bci_compet_iii}.

The UAM dataset uses only  
10 electrodes that are already known to perform well in P300 recognition problem, but
we can see from Figure \ref{fig:sel_sensor_uam} that the adaptive mixed-norm
regularizer further selects some sensors that are essentially
located in the occipital region. 
Note that despite the good
average performances reported in Table \ref{tab:perfresults},
some subjects in this dataset achieve  very poor performances, of about 50 \% of AUC, regardless of the considered method.
Selected channels for one of these subjects (Subject 25) are
depicted in Figure \ref{fig:sel_sensor_uam} and 
interestingly, they strongly differ from those of other subjects
providing rationales for the poor AUC.

\begin{figure}[t]
  \centering
  \includegraphics[width=0.9\columnwidth]{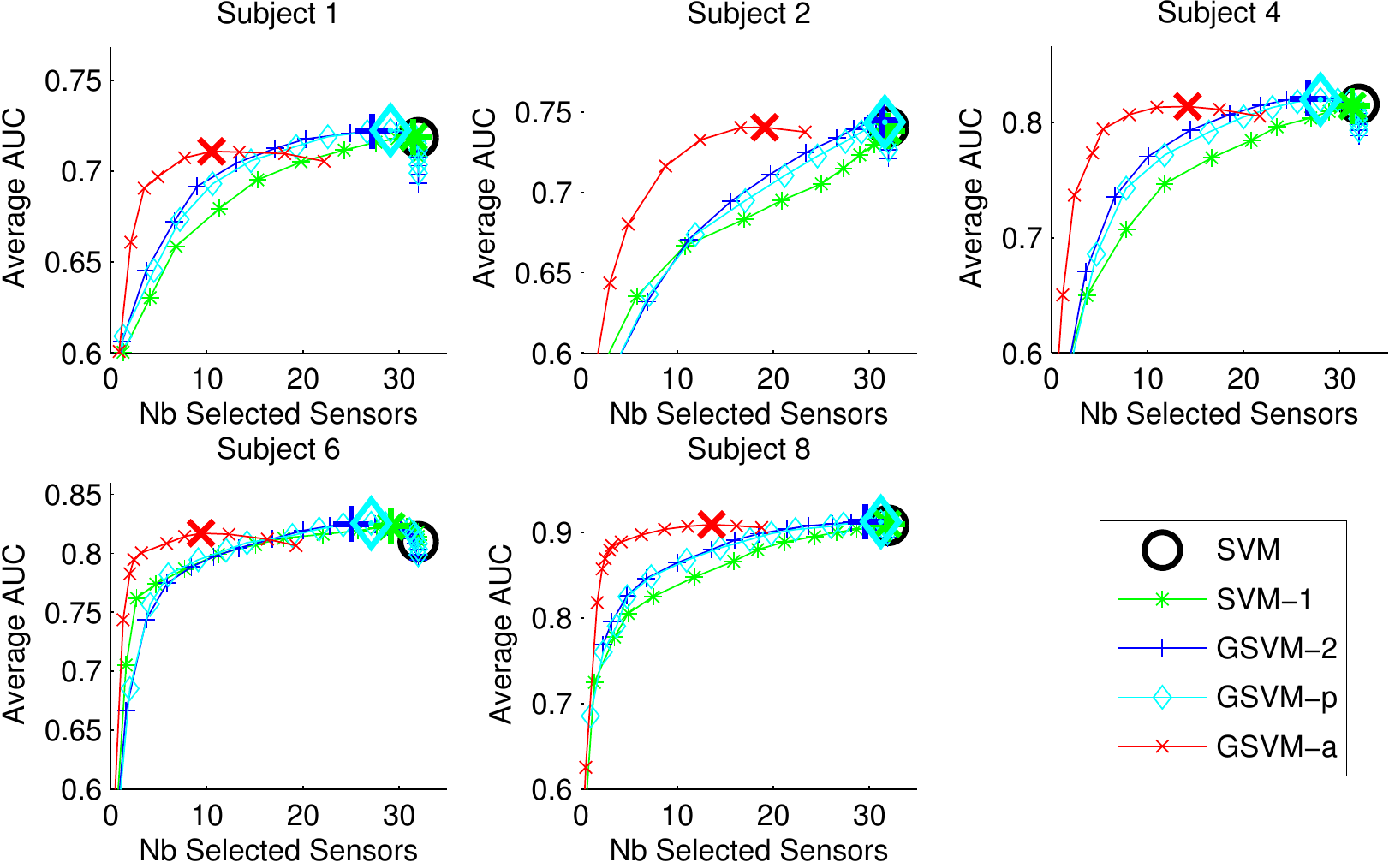}
  \caption{Performance vs sensor selection visualisation for the EPFL
    dataset. The large marker corresponds to the best model along
    the regularization path.}
  \label{fig:perf_svs_sparsity}
\end{figure}

\begin{figure}[t]
  \centering
  \includegraphics[width=0.9\columnwidth]{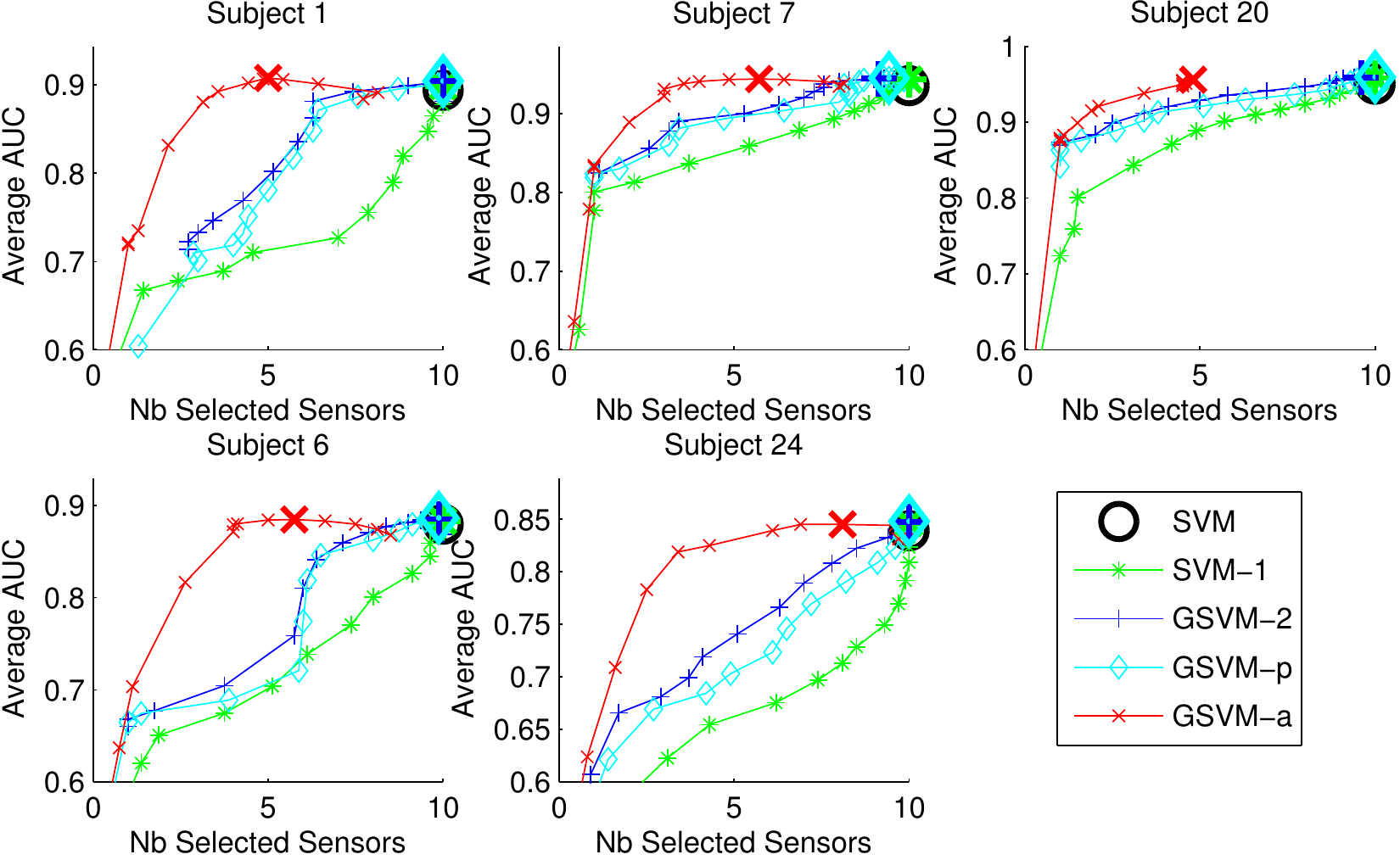}
  \caption{Performance vs sensor selection visualisation for the UAM
    dataset. The large marker corresponds to the best model along
    the regularization path. }
  \label{fig:perf_svs_sparsity_uam}
\end{figure}

We  have also investigated the impact of sparsity on the overall
performance of the classifiers. To this aim, we have plotted the average
performance of the different  classifiers as a function of the
number of selected sensors. These plots  are depicted in 
Figure~\ref{fig:perf_svs_sparsity} for the EPFL dataset and on
Figure~\ref{fig:perf_svs_sparsity_uam} for the UAM dataset. 
For both datasets, GSVM-a frequently achieves
a better AUC for a given level of sparsity. For most of the subjects, GSVM-a
performs as well as SVM but using far less sensors. 
A rationale may be that, in addition to selecting the
relevant sensors, GSVM-a may provide a better estimation of the classifier
parameters  leading to better performances for a fixed number of sensors.
As a summary, we suggest thus the use of an
adaptive mixed-norm regularizer  instead of
an $\ell_1-\ell_2$ mixed-norm as in van Gerven et
al. \cite{van2009interpreting}
when sparsity and channel selection is of primary importance.

\subsubsection{ErrP Dataset}
\label{sec:errp-dataset}

\begin{figure}[t]
  \centering
  \includegraphics[width=0.9\columnwidth]{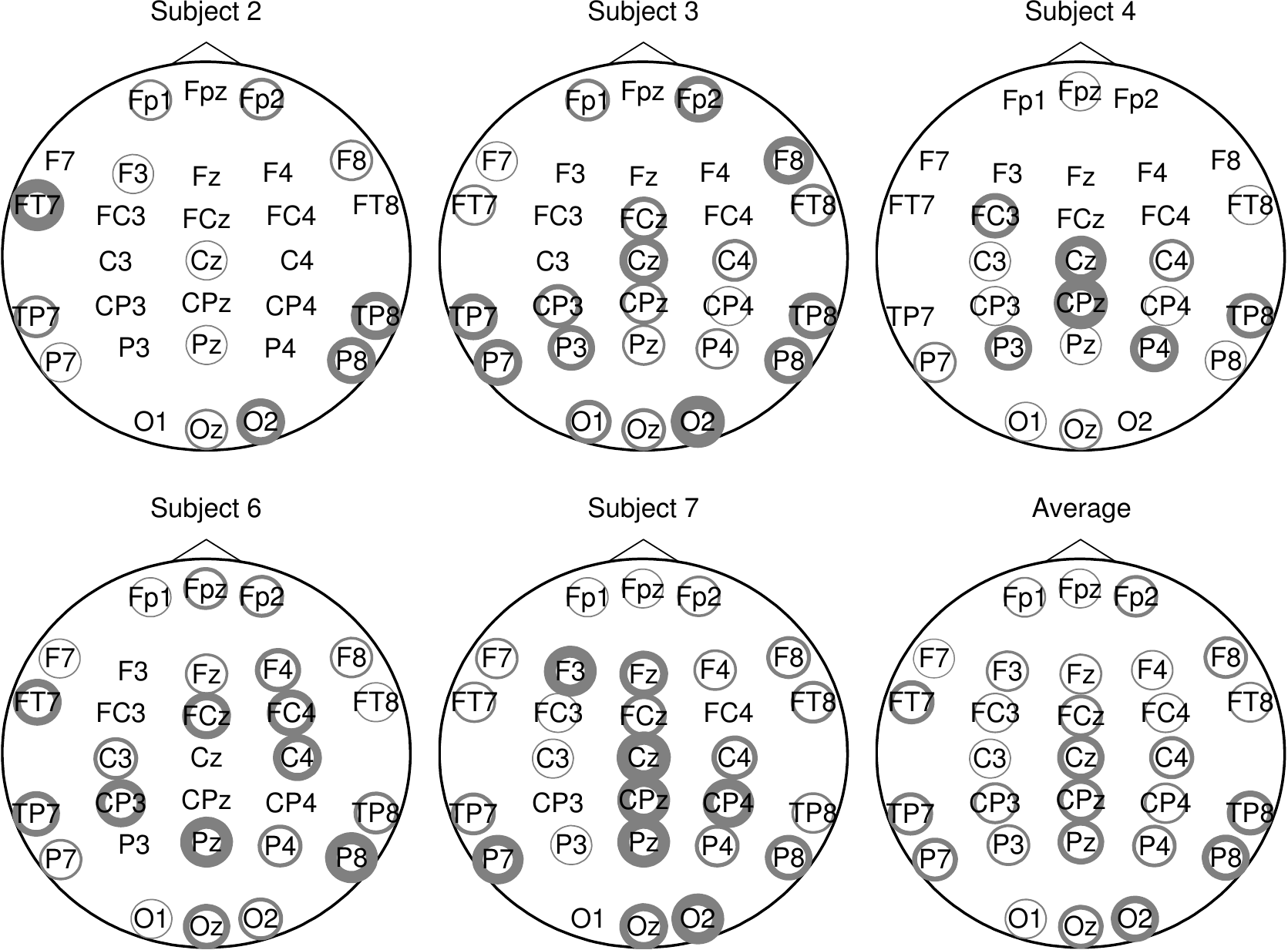}
  \caption{Selected Sensors for the ERP dataset. The line width of the
    circle is proportional to the number of times the sensor is
    selected. No circle means that the sensor has never been selected.}
  \label{fig:sel_sensor_erp}
\end{figure}

The ErrP dataset differs from the others as its number of examples is
 small (72 examples per subject).  The same experimental
protocol as above has been used for evaluating the methods but only 57
examples out of 72 have been retained for validation/training.
Classification performances are reported on Table
\ref{tab:perfresults}. For this dataset, the best performance is
achieved by GSVM-2 but the Wilcoxon test shows that all methods are
actually statistically equivalent. Interestingly, many channels of
this dataset seem to be irrelevant for the classification
task. Indeed, GSVM-2 selects only 30\% of them while GSVM-a uses only
7\% of the channels at the cost of 10\% AUC loss. We believe that this
loss is essentially caused by the aggressive regularization of GSVM-a and the
difficulty to select the regularization parameter $\lambda$ using only a subset  of the  
57 training examples.
Channels selected  by GSVM-2 can be visualized on Figure
\ref{fig:sel_sensor_erp}. Despite the high variance in terms of
selected sensors, probably due to the small number of examples,
sensors in the central area seem to be the most selected one,
which is consistent with previous results in
ErrP~\cite{dehaene1994localization}.

\subsubsection{Multi-task Learning}
\label{sec:multitask-learning}

We now evaluate the impact of the approach
we proposed in Equation (\ref{eq:mtlframework}) and
(\ref{eq:regterm_mtl}) on the 
P300 datasets.
 We expect that since
multi-task learning allows to transfer some information between the
different classification tasks, it will help in
leveraging classification
performances especially when the number of available training examples 
is small. Note that the ErrP dataset has not been tested
in this MTL framework, because the above-described results suggest an
important variance in the selected channels for all subjects. 
Hence, we believe that this learning problem does not fit into
the prior knowledge considered through Equation (\ref{eq:regterm_mtl}).

We have followed the same experimental protocol
as for the single task learning except that  
training and test sets have been formed as follows.
We first create training and test examples for a given subject 
 by randomly splitting  all examples of that subject, and
then  gather  all subject's training/test sets to form
the multi-task learning training/test sets. Hence, all the
subjects are equally represented in these sets.
A
$3$-fold nested cross-validation method is performed in order to automatically select the
regularization terms ($\lambda_r$ and $\lambda_s$).

\begin{figure}[t]
  \centering
    \includegraphics[width=0.47\columnwidth]{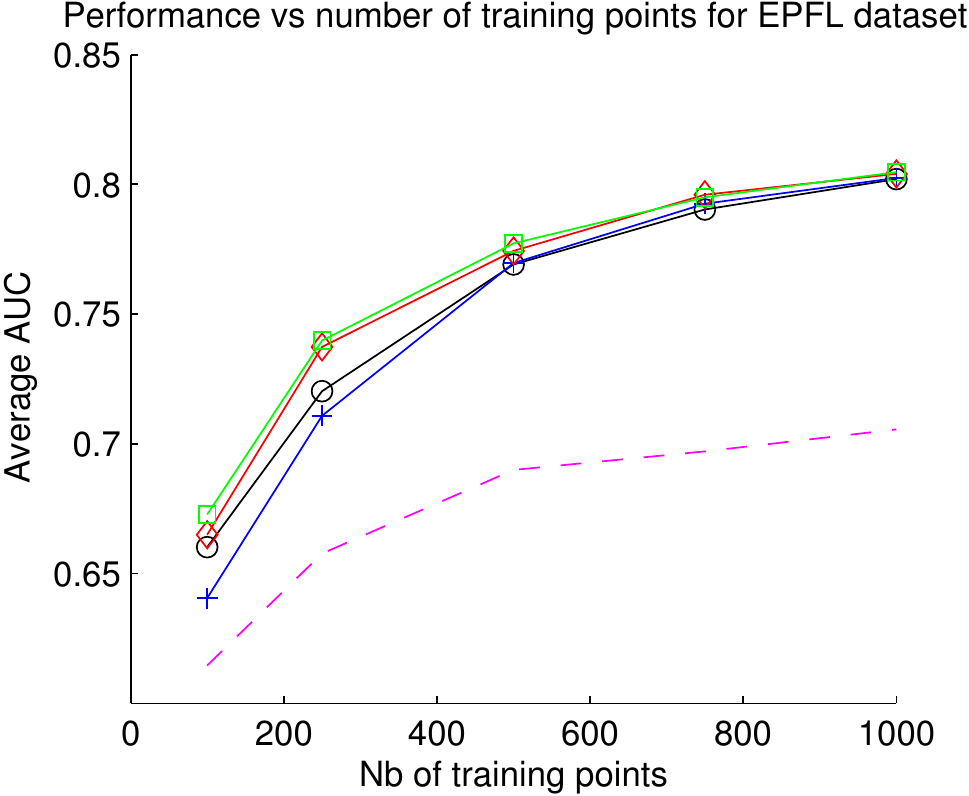}\label{fig:perf_mtl_epfl}
 \hspace{0.005cm}
 \includegraphics[width=0.47\columnwidth]{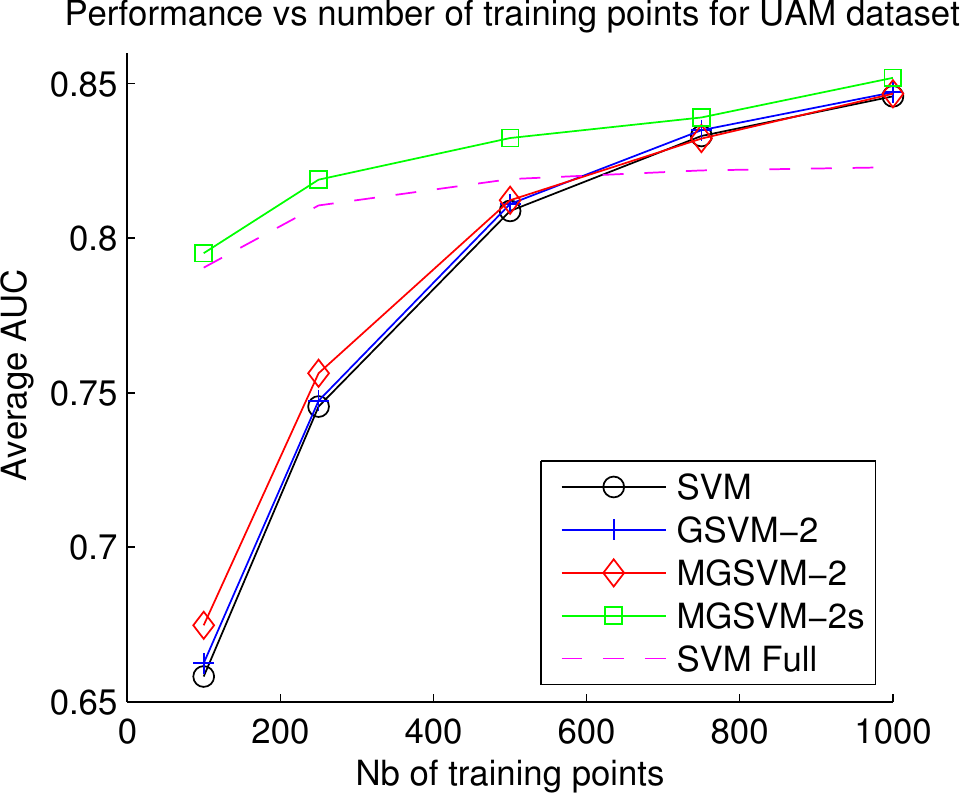}
  \caption{Multi-task learning performances (AUC) for the EPFL (left plot)
    and UAM (right plot) datasets for different number of training examples per subject.}
  \label{fig:mtl_epfl}
\end{figure}

\begin{figure}[t]
  \centering
  \includegraphics[width=0.45\columnwidth]{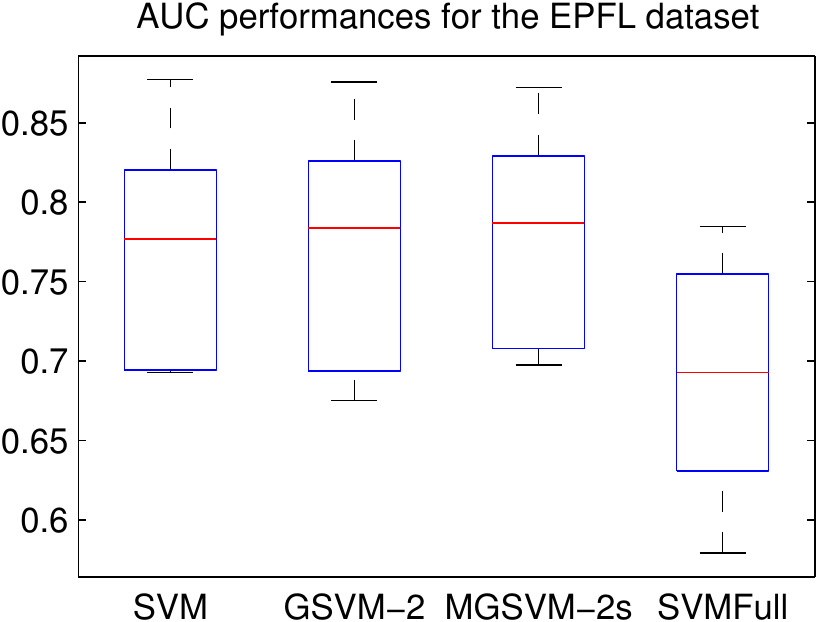}\hspace{3mm}
\includegraphics[width=0.45\columnwidth]{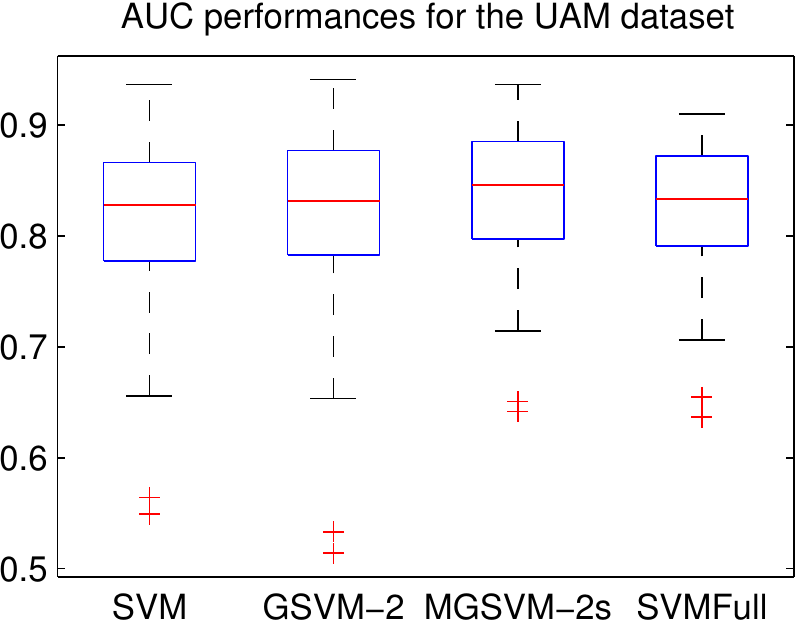}
  \caption{AUC performances comparison with EPFL (left plot) and UAM
    (right plot) for 500 training examples
    per subject.}
  \label{fig:aucperfmtl}
\end{figure}
Performances of the different methods have been  evaluated for
increasing number of training examples per subject and
are reported in Figure \ref{fig:mtl_epfl}. 
We can first see that for the EPFL dataset,  MGSVM-2 and
\mbox{MGSVM-2s} yield a slight but consistent improvement over the
single-task classifiers
(SVM-Full being a single classifier trained on all subject's examples
and SVM being the average performances of subject-specific classifiers).
The poor performances of the SVM-Full approach is  probably due 
to the high inter-subject variability in this dataset, which includes
impaired patients. 

For the UAM dataset,  results are quite
different since the SVM-Full and \mbox{MGSVM-2s} shows a
significant improvement
over the single-task learning. 
We also note that, when only the joint channel
selection regularizer is in play (\mbox{MGSVM-2}),  multi-task learning
leads to poorer performance than the SVM-Full for a number
of trials lower than $500$. We  justify this by 
the difficulty of achieving appropriate
channel selection based only on few training examples, as
confirmed by the performance of GSVM-2.  
\cor{
From  Figure \ref{fig:aucperfmtl}, we can see that
the good performance of  \mbox{MGSVM-2s}  is the outcome of 
 performance improvement of about 10\% AUC over SVM, 
 achieved on some subjects that perform
poorly. More importantly, while performances of these
subjects are significantly increased, those that performs
well still achieve good AUC scores. In addition, we emphasize that
these improvements are essentially due to the similarity-inducing
regularizer. 
}

For both datasets, the MTL approach \mbox{MGSVM-2s} is consistently
better than those of other single-task approaches thanks to the
regularization parameters $\lambda_r$ and
$\lambda_s$ that can adapt to the inter-subject similarity (weak similarity for
EPFL and strong similarity for UAM).
These are
interesting results showing that multi-task learning can be a way to
handle the problem related to some subjects that achieve
poor performances. Moreover, results also indicate that
multi-task learning is useful for drastically shortening the
calibration time. For instance, for the UAM dataset, 80\% AUC was
achieved using only 100 training
examples (less than 1 minute of training example recordings).
Note that the validation procedure tends to maximize
  performances, and does not lead to sparse classifiers for MTL
  approaches. As shown in Figures \ref{fig:sel_sensor_epfl} and
  \ref{fig:sel_sensor_uam}, the relevant sensors are quite different
  between subjects thus a joint sensor selection can lead to a slight
  loss of performances, hence   the tendency of the cross-validation  procedure to select non-sparse
classifiers.

\section{Conclusion}
\label{sec:conclusion}

In this work, we have investigated the use of mixed-norm regularizers 
for  discriminating Event-Related Potentials in BCI.  We have 
extended the discriminative framework of van Gerven et al.
\cite{van2009interpreting} by studying general mixed-norms and
proposed the use of the adaptive mixed-norms as sparsity-inducing
regularizers.  This discriminative framework has
been broadened to the
multi-task learning framework where  classifiers
related to different subjects are jointly trained. For this
framework, we have introduced a novel regularizer that induces
channel selection and classifier similarities.
The different proposed approaches were tested on three different datasets
involving a substantial number of subjects. Results from 
these experiments have highlighted that the 
$\ell_1-\ell_2$ regularizer 
has been proven interesting for improving classification performance whereas 
adaptive mixed-norm is the regularizer to be considered when sensor
selection is the primary objective.
Regarding the multi-task learning framework, 
our most interesting finding is that this learning framework
allows, by learning more robust classifiers, significant performance improvement on some  subjects 
that perform poorly in a single-task learning context.    

\remi{}{In future work, we plan to investigate a different grouping
  of the features, such as temporal groups. This kind of group
  regularization could be for instance used in conjunction with the
  sensors group in order to promote both feature selection and
  temporal selection in the classifier. While the resulting problem is still
convex, its resolution  poses some issues so that a dedicated solver
would be necessary.

Another research direction would be to investigate the use of
asymmetrical MTL. This could prove handy when a poorly-performing subject  will negatively influence the other subject performances in MTL while
improving his own performances. In this
case one would like that subject classifier to be similar to the
other's classifier without impacting their classifiers. }

\addcontentsline{toc}{section}{References}
\bibliographystyle{elsart-harv}

\section*{Appendix}
 \subsection{Proof of Lipschitz gradient of the squared Hinge loss}
\label{sec:append_lip}
Given the training examples $\{\x_i,y_i\}$, the squared Hinge loss
is written as :
$$
J=\sum_{i=1}^n \max(0,1-y_i \x_i^\top \w)^2
$$
and its gradient  is :
$$
\nabla_\w J=-2\sum_{i} \x_iy_i\max(0,1-y_i \x_i^\top \w)
$$
 The squared Hinge loss is gradient Lipschitz if there exists a constant $L$ such that:
$$
\|\nabla J(\w_1) - \nabla J(\w_2)\|_2 \leq L \|\w_1 - \w_2\|_2 
\quad \forall \w_1,\w_2 \in \R^d.
$$
The proof essentially relies on showing that $\x_iy_i\max(0,1-y_i \x_i^\top \w)$
is Lipschitz itself \emph{i.e} there exists $L^\prime \in \R$ such that
\begin{align}\nonumber
&\|\x_iy_i \max(0,1-y_i \x_i^\top \w_1) - \x_iy_i \max(0,1-y_i \x_i^\top \w_2)  \| \\ &
\leq L^\prime \|\w_1 - \w_2\| \nonumber
\end{align}
Now let us consider different situations. For a given 
$\w_1$ and $\w_2$, if 
  $1-\x_i^T\w_1 \leq 0$
and   $1-\x_i^T\w_2 \leq 0$, then the left hand side is equal to $0$
and any $L^\prime$ would satisfy the inequality.
 If 
  $1-\x_i^T\w_1 \leq 0$
and   $1-\x_i^T\w_2 \geq 0$, then the left hand side (lhs) is
\begin{eqnarray}
  \label{eq:lipschitz}
 lhs &=& \|\x_i\|_2 (1-\x_i^\top \w_2) \\ \nonumber
&\leq& \|\x_i\|_2(\x_i^\top\w_1 -\x_i^\top \w_2)\\ \nonumber
&\leq& \|\x_i\|_2^2 \|\w_1-\w_2\|_2 \nonumber
\end{eqnarray}
A similar reasoning yields to the same bound when 
 $1-\x_i^T\w_1 \geq 0$ $1-\x_i^T\w_1 \leq 0$
and   $1-\x_i^T\w_2 \geq 0$
and   $1-\x_i^T\w_2 \geq 0$. Thus,  $\x_iy_i\max(0,1-y_i \x_i^\top \w)$ is Lipschitz
with a constant $\|\x_i\|^2$. Now, we can conclude
the proof by stating that $\nabla_\w J$ is Lipschitz
as it is a sum of Lipschitz function and
the related constant is $\sum_{i=1}^n\|\x_i\|_2^2$.  

\subsection{Lipschitz gradient for the multi-task learning problem}
\label{sec:append_lip2}
For the multi-task learning problem, we want to prove
that the function
$$
\sum_{t=1}^m \sum_{i=1}^{n} L(y_{i,t},\x_{i,t}^\top \w_t + \b_t)
+ \lambda_s \sum_{t=1}^m \|\w_t - \frac{1}{m}\sum_{j=1}^m \w_j\|_2^2
$$
is gradient Lipschitz, $L(\cdot,\cdot)$ being the square Hinge loss.
From the above results, it is easy to show that the first term is
gradient Lipschitz as the sum of gradient Lipschitz functions. 

Now, we  also show that the similarity term 
$$\sum_t \|\w_t - \frac{1}{m}\sum_{j=1}^m \w_j\|_2^2$$ is also gradient Lipschitz.

This term can be expressed as
\begin{eqnarray}\nonumber
 \|\w_t - \frac{1}{m}\sum_{j=1}^m \w_j\|_2^2&=& \sum_{t} \langle \w_t,\w_t \rangle
- \frac{1}{m} \sum_{i,j=1}^m \langle \w_i,\w_j \rangle \\ \nonumber
&=&\w^\top \M \w \nonumber
\end{eqnarray}
 where
 $\w^\top=[\w_1^\top, \dots,  \w_m^\top]$ is the vector of all
 classifier parameters and $\M\in\dbR^{md\times md}$ 
is the Hessian
 matrix of the similarity regularizer of the form
$$\M=\I-\frac{1}{m}\sum_{t=1}^m\D_t$$
with $\I$ the identity matrix and $\D_t$ a block 
matrix with $\D_t$ a $(t-1)$-diagonal matrix where
each block is an identity matrix $\I$ with appropriate
circular shift. $\D_t$ is thus a $(t-1)$ row-shifted version
of $\I$.

Once we have this formulation, we can use the fact that
a function $f$ is gradient Lipschitz  of constant $L$ if the
largest eigenvalue of its Hessian  is bounded by $L$
on its domain \cite{bertsekas_nonlinear}. 
Hence, since we have 
$$
  \|\M\|_2\leq \|\I\|_2+\frac{1}{m}\sum_{t=1}^m\|\D_t\|_2=2
$$
the Hessian matrix of the similarity term $2 \cdot \M$ has
consequently bounded eigenvalues. This concludes the proof that the
function $\w^\top\M\w$ is gradient Lipschitz continuous.

\subsection{Proximal operators}
\label{sec:append_prox}
 \subsubsection{$\ell_1$ norm}

 the proximal operator of the $\ell_1$ norm is defined as :
 $$
 \text{prox}_{\lambda \|\x\|_1} (\mathbf{u}) = \arg \min_{\x} \frac{1}{2} \| \x - \u \|_2^2+
 \lambda \|\x\|_1
 $$
 and has the following closed-form solution for which each component
 is
 $$
 [ \text{prox}_{\lambda \|\x\|_1} (\mathbf{u})]_i  = \text{sign}(u_i) (|u_i| - \lambda)_+
 $$
 \subsubsection{$\ell_1-\ell_2$ norm}
the proximal operator of the $\ell_1-\ell_2$ norm is defined as :
 $$
 \text{prox}_{\lambda \sum_{g\in \mathcal{G}}\|\x_g\|_2} (\mathbf{u}) = \arg \min_{\x} \frac{1}{2} \| \x - \u \|_2^2+
\lambda \sum_{g\in \mathcal{G}}\|\x_g\|_2
 $$
the minimization problem can be decomposed into several ones since the
indices $g$ are separable. Hence, we can just focus on the problem
$$
\min_{\x} \frac{1}{2} \| \x - \u \|_2^2+
\lambda \|\x\|_2
$$
which minimizer is 
$$
\left\{
\begin{array}{ll}
0 & \text{ if } \|\mathbf{u}\|_2 \leq \lambda \\
(1-\frac{\lambda}{\|\mathbf{u}\|_2}) \mathbf{u} & \text{otherwise}
\end{array}\right.
$$


  \end{document}